\documentclass[11pt,a4paper]{article}
\usepackage[hyperref]{eacl2021}
\usepackage{times}
\usepackage{latexsym}

\usepackage{microtype}
\usepackage{float}
\aclfinalcopy 
\def\aclpaperid{821} 

\usepackage{graphicx}
\usepackage{gb4e} 
\noautomath  
\usepackage{caption}
\usepackage{subcaption}
\usepackage{multicol}

\usepackage{ragged2e}
\AtBeginEnvironment{multicols}{\RaggedRight}

\definecolor{plot_blue}{rgb}{0.0, 0.2, 0.9}
\definecolor{capri}{rgb}{0.0, 0.70196, 0.649}
\definecolor{AbC_color}{rgb}{0.75, 0, 0.75}

\usepackage[colorinlistoftodos]{todonotes}
\usepackage{lipsum}

\title{Language Modelling as a Multi-Task Problem}

\author{Lucas Weber \\
        DTCL, University Pompeu Fabra\\
        \texttt{lucas.weber@upf.edu} \\
        \And
        Jaap Jumelet \\
        ILLC, University of Amsterdam\\
        \texttt{j.w.d.jumelet@uva.nl} \\
        \AND
        Elia Bruni \\
        IKW, University of Osnabr{\"u}ck\\
        \texttt{elia.bruni@gmail.com} \\
        \And
        Dieuwke Hupkes \\
        Facebook AI Research\\
        \texttt{dieuwkehupkes@fb.com} \\}

\date{}

\begin{document}

\maketitle

\begin{abstract}
In this paper, we propose to study language modelling as a multi-task problem, bringing together three strands of research: multi-task learning, linguistics, and interpretability.
Based on hypotheses derived from linguistic theory, we investigate whether language models adhere to learning principles of multi-task learning during training.
To showcase the idea, we analyse the generalisation behaviour of language models as they learn the linguistic concept of Negative Polarity Items (NPIs). 
Our experiments demonstrate that a multi-task setting naturally emerges \emph{within} the objective of the more general task of language modelling.
We argue that this insight is valuable for multi-task learning, linguistics and interpretability research and can lead to exciting new findings in all three domains.
\end{abstract}

\section{Introduction}
\label{sec:introduction}

\begin{figure*}
    \centering
    \includegraphics[width=0.7\linewidth, trim=0mm 0mm 0mm 0mm, clip]{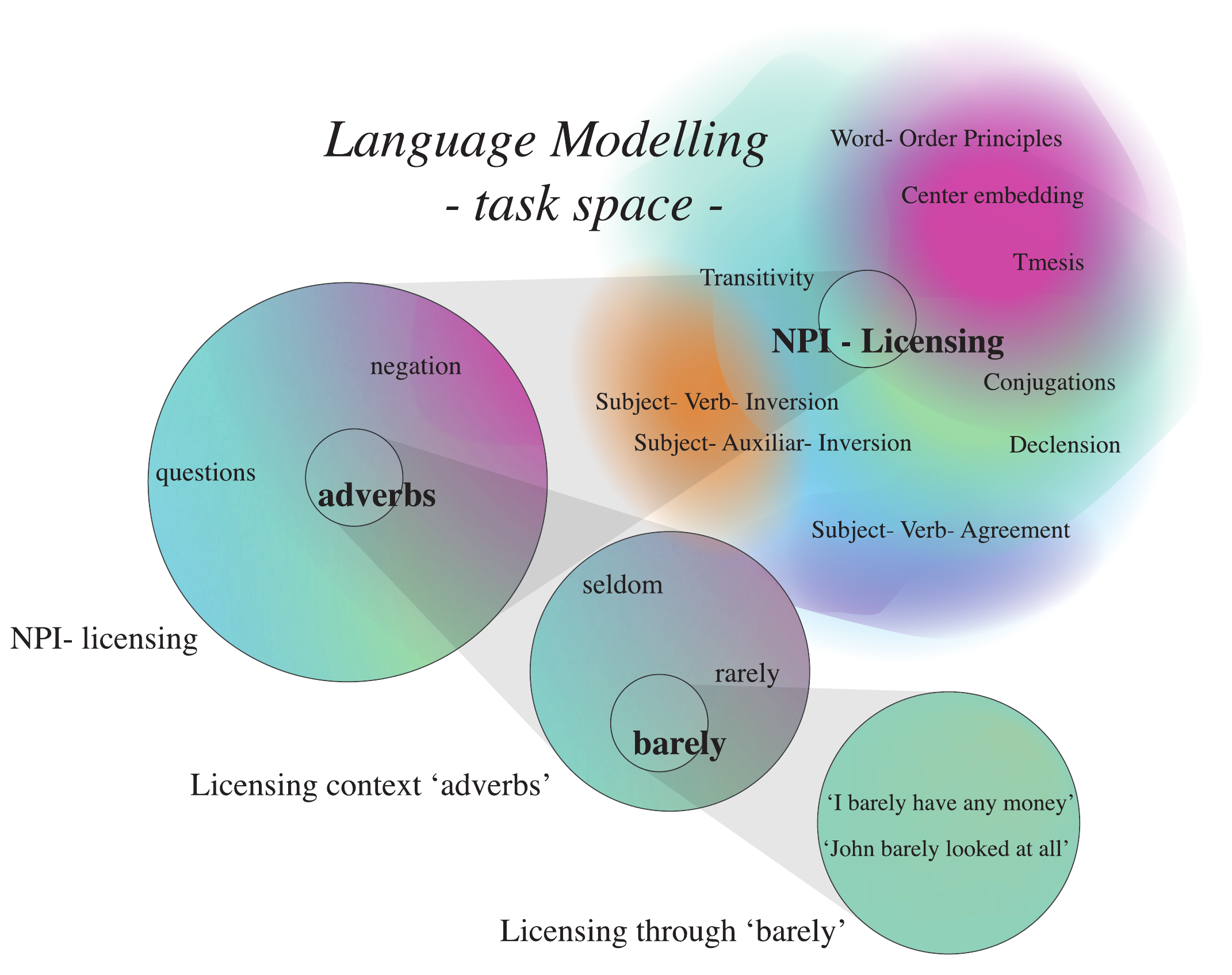}
    \caption{A conceptual visualisation of a language modelling task hierarchy, from language modelling as a whole to single examples, with complex similarities between tasks. Colours indicate task similarities.}
    \label{fig:task_hierarchy}
\end{figure*}

Humans are optimising their behaviour towards a multitude of objectives to reach their goals in day-to-day life. 
By learning many things at the same time and exploiting their commonalities, they acquire more general knowledge about the world, which in turn helps them to learn new things quicker \citep{perkins1992transfer, schwartz2005efficiency, cormier2014transfer, luria1976cognitive}.
This idea of finding more general solutions through the diversification of tasks has found its way also to the machine learning community, in the field of multi-task learning (MTL) \citep{Caruana93multitasklearning:,Caruana1997MultitaskLearning}. 
In MTL, multiple tasks are optimised jointly, enabling the transfer of relevant information across tasks. 
MTL research yields fruitful results in both application \citep[e.g.][]{Collobert2008ALearning, Collobert2011NaturalScratch,zhang2014facial, donahue2014decaf, kaiser2017model} and theory \citep[e.g.][]{baxter2000model, maurer2006bounds, ando2005framework, argyriou2008spectral}. 

However, deciding on a setup requires making many arbitrary choices.
The researcher or engineer has to decide which tasks to train together \citep[e.g.][]{bingel-sogaard-2017-identifying, Standley2019WhichTS}; at which hierarchy-level to allow tasks to interact \citep[e.g.][]{sogaard-goldberg-2016-deep}; which degree of parameter sharing to employ \citep{Ruder2017AnNetworks}; which distribution of training data to employ \citep[e.g.][]{Luong2016MultitaskST}, and so on.
Having to make so many arbitrary choices is inconvenient for modellers, but also stands in the way of understanding the learning principles of neural models in multi-task settings. 
The highly constructed learning scenarios make it difficult to see whether outcomes should be attributed to one of the many a-priori decisions or to inherent properties of the learning process.

In this paper, we propose to study MTL not in a constructed, artificial scenario, but in a more natural setting. 
To do so, we consider the objective of \emph{language modelling} and exploit the fact that it can be seen as a conglomerate of many different tasks.
To give an example: rules of word ordering have to be learned simultaneously to rules of feature agreement and the monotonicity properties of different linguistic environments.
These different tasks all need to be learned to achieve the greater goal of producing acceptable sentences, and they have to be optimised in parallel when the language model (LM) is trained.
Language modelling is in that sense a \textit{natural} multi-task learning problem with a naturally given \emph{task hierarchy} provided by linguistic theory (see also Figure~\ref{fig:task_hierarchy}). 

Studying language modelling as a multi-task problem has several distinct advantages. 
From an MTL perspective, it gives us a complete hierarchy of relevant tasks that can freely interact throughout the learning process, unconstrained by prior assumptions.
We can make theoretically informed decisions about these tasks, drawing on linguistic theory.
We can also deduce from linguistics how these tasks relate to each other (or, in other words, how similar they are), which in MTL is considered to be one of the crucial factors for the learning outcomes \citep[e.g.][]{Thrun1996DiscoveringAlgorithm, Passos2012FlexibleLearning}. 
MTL has not yet been studied from this dynamic and unconstrained perspective.
Then, somewhat more delicately, the extent to which models can exploit similarities hypothesised by linguistic theory can play a role in confirming or refuting specific linguistic hypotheses.
Lastly, when it comes to interpretability research, applying concepts from MTL can be valuable to better understand the learning dynamics of models.
By understanding how models are finding solutions, we can infer what these solutions are.

\paragraph{Outline} In the remainder of this paper, we will first provide some basic background about MTL (\S~\ref{subsec:multi_task_learning}), the subset of linguistic tasks we focus on (\emph{Negative Polarity Items}, where we consider their different \emph{licensing contexts} as tasks, \S~\ref{subsec:npis}) and discuss some related work in interpretability (\S~\ref{subsec:interpretability}).
Then, in \S~\ref{sec:approach} and \S~\ref{sec:experiments}, respectively, we present our approach and empirical results that showcase our idea.
In \S~\ref{sec:discussion}, we discuss our results and framework in the light of the three fields mentioned before.
We conclude in \S~\ref{sec:conclusion}.

\section{Background}
\label{sec:background_related_work}

In this paper, we aim to bring together three strands of research: MTL, linguistics and interpretability research.
As a proof of concept, we focus on one specific complex subset of linguistic tasks: \textit{licensing of Negative Polarity Items (NPIs)}. 
Below, we give a short overview of the most important characteristics of the three fields of interest.

\subsection{Multi-task learning}
\label{subsec:multi_task_learning}

In MTL, multiple tasks are learned together to enable information transfer from one task to another.
If the transfer is successful, the benefits might be threefold: the model learns tasks with less training data \citep[i.e.\ \textit{more efficient},][]{Collobert2011NaturalScratch, benton-etal-2017-multitask, kaiser2017model}, up to a higher final accuracy \citep{Collobert2008ALearning,kaiser2017model} and in a way that better generalises to new tasks \citep{baxter2000model, Collobert2008ALearning}.

\citet{Caruana93multitasklearning:,Caruana1997MultitaskLearning} and \citet{Ruder2017AnNetworks} propose several different -- but related -- processes that might enable positive transfer: related tasks can provide additional training examples for each other on the features they share (\emph{statistical data amplification}), certain features might be easier to learn through one task than through another, but be useful for both of them (\textit{eavesdropping}), and idiosyncratic features of single tasks can be averaged out, while more general features are reinforced (\textit{attention focusing})\footnote{For a complete list of processes please consult the original publications.}. 

However, positive transfer is not guaranteed; It is also possible that performance \emph{deteriorates} due to interference between different tasks, resulting in negative transfer, \cite{Rosenstein05totransfer, Pan2010survey, Wang2019negativetransfer}. 
Whether transfer is positive depends on the \textit{task similarity} and whether the model is able to exploit this similarity  \citep{Rosenstein05totransfer, Thrun1996DiscoveringAlgorithm, Passos2012FlexibleLearning}.

The main goal of MTL so far has been to avoid negative- and promote positive transfer by determining task-similarity and regulate the interactions between tasks based on these similarities. Due to its pivotal role, much research effort was spent on determining similarities of tasks and the regulation of information transfer between them \citep[for an overview, see][]{Zhang2017ALearning,Ruder2017AnNetworks}.
The disadvantage of these approaches is that assuming fixed tasks and regulating transfer between them based on fixed task-similarities puts large constraints on possible transfers between tasks, because it neglects the fact that learning processes are dynamic.
From the perspective of the model, tasks, as well as their similarities, can change throughout the learning process.
Here, we only use predefined tasks and their similarities to \textit{analyse} the learning behaviour of the model, without constraining the learning process in any way.


\subsection{Negative Polarity Items}
\label{subsec:npis}
We exemplify our idea by analysing the learning behaviour on a complex subset of linguistic tasks: the licensing of Negative Polarity Items (NPIs).
The properties of NPI licensing make it an interesting and adequate subset of tasks to study, as it has a high degree of complexity, has an appropriate frequency within natural language and was previously frequently investigated in neural models.


NPIs are characterised by the property that they can only occur within the scope of certain \textit{licensing contexts}.
For instance, in the example below, the NPI `\textbf{any}' can occur in sentence (1)a., where it is in the scope of a negation, but not in sentence (1)b., where there is no licensor present.
\begin{exe}
    \ex\begin{xlist}
    \ex[]{Bill did\textit{n't} buy \textbf{any} books that day.}
    \ex[*]{Bill did buy \textbf{any} books that day.}
    \end{xlist}
    
    \ex\begin{xlist}
    \ex[]{\textit{Nobody} has \textbf{ever} been there.}
    \ex[*]{Somebody has \textbf{ever} been there.}
    \end{xlist}
\end{exe}

Licensing contexts are formed on the basis of semantic properties, such as downward entailment \citep{Fauconnier1975, Ladusaw1980PolarityRelations}, non-veridicality \citep{Giannakidou2011NegativeCompositionality}, or scope marking \citep{Barker2018}.
Common licensing contexts include negation, conditionals, or superlatives, and are often \textit{triggered} by a specific expression, such as `not' or `nobody'.

Grasping the phenomenon of NPI licensing requires understanding of three different aspects:

\begin{enumerate}\setlength{\itemsep}{0mm}
    \item \textit{The class of NPIs}: there is a group of expressions that are restricted in their occurrence.
    \item \textit{Licensing contexts}: there exists a group of expressions that allow NPIs to occur.
    \item \textit{Scope and structure}: the licensing contexts have to stand in a certain structural relationship to the NPIs.
\end{enumerate} 

We focus on how LMs learn the second aspect by analysing how different types of licensing contexts interact and generalize throughout training.
During learning they should be able to exploit their similarity in the other two aspects.

\subsection{Interpretability}\label{subsec:interpretability}

Interpretability research on LMs has shown that in pre-trained models, such as BERT \citep{devlin-etal-2019-bert}, hierarchical structure emerges throughout the layers and that this structure demonstrates parallels with linguistic theory \citep{peters2018dissecting,liu2019linguistic,tenney-etal-2019-bert}. 
However, the emergence of this structure has not been explicitly connected to MTL yet.

In recent years, research has shown that LMs are able to understand NPI licensing.
\citet{jumelet-hupkes-2018-language} evaluate the performance of LMs on data sets containing NPI constructions extracted from large corpora, and \citet{Marvin2018TargetedModels, Wilcox2019, warstadt2020can} test them on artificial data sets containing template-based NPI constructions.
In our own experimental setup we will utilise the extensive template-based NPI corpus of \citet{Warstadt2019InvestigatingNPIs}.

What these approaches have in common is their focus on the performance of pretrained LMs.
Our MTL approach sheds light on an unexplored aspect of NPI understanding: the learning dynamics of the model \textit{during} training.

\section{Approach}
\label{sec:approach}


We consider two different types of experiments.
First, to understand to which extent models can understand and use the similarity between different licensing contexts (our \emph{tasks}) during learning, we exploit the effect that frequency of the different contexts has on learning.
Second, we manipulate the LMs' training corpus to constrain their ability to leverage information from other licensing contexts during learning.
In accordance with the MTL-literature, we expect the LMs to learn tasks more data-efficient and to a higher final accuracy if they can leverage information across contexts.
Before we describe our experiments in more detail, we present our model architecture and training, the evaluation procedure of the licensing contexts, and the filter procedure we use to manipulate the training corpus.

%

\subsection{Model}
\label{architecture_and_training}
Following previous work in this area, we consider recurrent language models. 
We focus on uni-directional LSTM models and mirror the hyperparameter setup of \citet{gulordava-etal-2018-colorless}\footnote{Hyperparameters: batch size = 64, BPTT length = 35, dropout = 0.1, adaptive SGD learning rate = 20, layers = 2, hidden and embedding size = 650, epochs = 40.}.
We train the models on the corpus provided by the same authors\footnote{\url{https://github.com/facebookresearch/colorlessgreenRNNs/tree/master/data}} -- a subset of the English Wikipedia -- or modified versions of the same for our second experiment (see \S~\ref{exp_2}).
To track the learning process, we save models every 100 batches of training (371 model-checkpoints per epoch). 
For all experiments, we average performance across five random seeds.

\subsection{Evaluation} 
\label{model_evaluation}

\begin{table*} 
    \small
    \renewcommand{\arraystretch}{1.3}
        \begin{tabular}{lp{0.6\linewidth}c} 
            \hline
             &  & \multicolumn{1}{l}{\textbf{Frequency} per}\\
            \textbf{Context} & \textbf{Example} & 100k sentences  \\ 
            \hline 
             Simple Questions & Did he \textbf{ever} do a mean thing\textit{?} & 10 \\
             Adverbs   & In the present political culture, there are \textit{hardly} \textbf{any} leaders who would avoid limelight and refuse positions of power.  & 23 \\ 
             Questions & However, various writers attribute it to Putnam, Stark, Prescott or Gridley, while others question \textit{whether} it was said \textbf{at all} .   & 25 \\ 
             Superlative & [...] and caused the \textit{worst} winter flooding \textbf{in decades} for river and stream valleys [...].  & 32 \\ 
             Only    & [...] "Those [students] \textit{only} are supposed to pay \textbf{anything} who are abundantly able, or prefer to do so. & 85 \\ 
             Conditional &  In 1997 Li published a paper attempting to replicate \texttt{<unk>}'s results and showed the effect was very small, \textit{if} it existed \textbf{at all}. & 127 \\ 
             Quantifier &   That's \textit{all} you'll \textbf{ever} need. & 179 \\ 
             Determiner negation & In spite of the \texttt{<unk>} of the disaster, \textit{no} one was \textbf{ever} held accountable. & 218 \\ 
             Sentential negation & It is \textit{not} judged under \textbf{any} subjective points of view, only the clock. & 712 \\ 
            \hline      
        \end{tabular}
    \caption{
     The nine types of licensing contexts taken from \citet{Warstadt2019InvestigatingNPIs}, with an example and the context frequency within the training corpus.
    }\label{table:warstadt}
\end{table*}

To estimate the LMs' understanding of NPIs and their dependence on the different licensing contexts, we adapt the Cloze task of \citet{Warstadt2019InvestigatingNPIs}, based on the implementation of \citet{jumelet-2020-diagnnose}.  
This task considers nine different types of licensing contexts (a list of the contexts, including examples, can be found in Table~\ref{table:warstadt}). 
For every such context, \citet{Warstadt2019InvestigatingNPIs} generated a large number of \emph{minimal pair sentences}, containing correctly and incorrectly licensed NPIs.
For instance, for the \emph{adverbs} licensing context:
\begin{exe}
    \ex\begin{xlist}
        \ex[]{A lady \textit{rarely} \textbf{ever} thought that the children saw the boy.}
        \ex[*]{A lady \textit{sometimes} \textbf{ever} thought that the children saw the boy.}
    \end{xlist}
\end{exe}
 
\begin{figure}[H]
\includegraphics[width=\columnwidth,trim={0 4cm 0 4cm},clip]{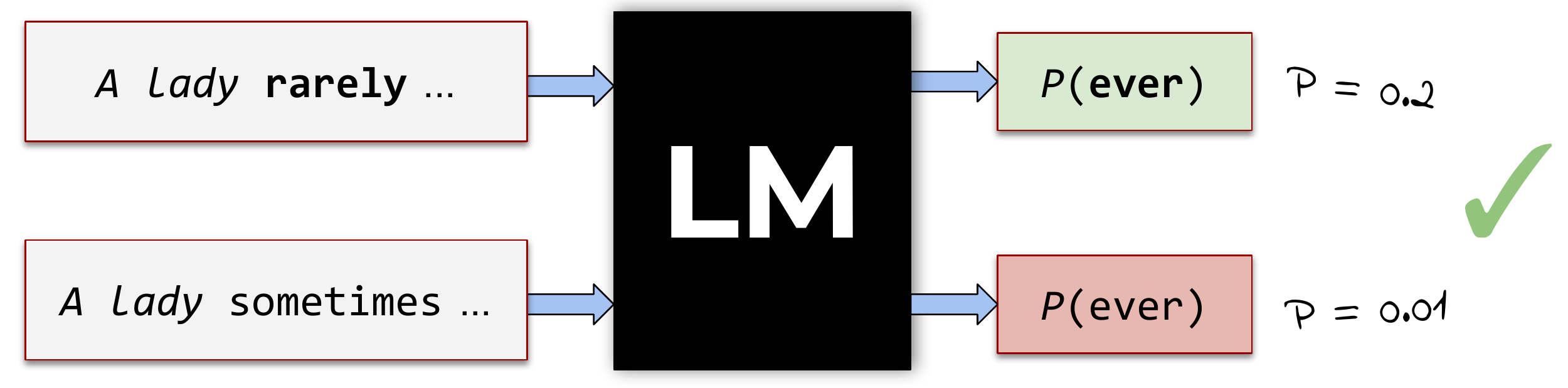}
 \caption{The NPI judgement task that is used for evaluating the LMs.
 A correct prediction assigns a higher probability to an NPI in a context that licenses it, based on the corpus of \citet{Warstadt2019InvestigatingNPIs}.
 }
 \label{fig:warstadt_task}
\end{figure}

Following previous work, we quantify an LM's understanding of a particular type of licensing context by computing the percentage of minimal pairs in that context for which the model correctly assigns a higher probability to the NPI in the licensing contexts than in the non-licensing contexts.
I.e., in the example above, we would compare the probability the model assigns to the word \emph{ever} in the contexts ``A lady rarely" and ``A lady sometimes" (see also Figure~\ref{fig:warstadt_task}).

\subsection{Identification of NPIs in training corpus}
\label{filtering}
The \citet{Warstadt2019InvestigatingNPIs} corpus provides us with a task to evaluate nine different context types that license NPIs.
To manipulate the training corpus for our experiments we also need to identify sentences in the training corpus of the model in which these contexts actually licence NPIs.
To do so, we need to locate these contexts, as well as establish that they in fact licence an NPI in a particular sentence.

We consider the nine \citeauthor{Warstadt2019InvestigatingNPIs} context types, and the corresponding list of 30 expressions that are part of these contexts (e.g.\ the list of adverbs licensing NPIs).
As for the NPIs, we consider an extensive list of 160 distinct NPIs\footnote{This list can be found in Appendix~\ref{sec:appendix_2}.}, based on the collection provided by \citet{Hoeksema2012OnProject}.
We then identify sentences in which an element of our NPI list is preceded by an element from our context list, ensuring that there is a dependency relation between them using the dependency parser of spaCy \citep{honnibal-johnson:2015:EMNLP}.
When there are multiple potential licensors in a sentence, we use the hierarchical distance between the licensor and the NPI in the parse tree as a heuristic to find the correct licensor. 
By testing this procedure on a manually labeled set of 200 randomly selected sentences with multiple licensors, we estimate that it identifies the correct among multiple licensors in around 97\% of cases.
%
%
In Table~\ref{table:warstadt}, we report examples and frequencies of the different licensing contexts in the training corpus based on this filtering scheme.

\section{Experiments and results}
\label{sec:experiments}
\begin{figure}
     \includegraphics[width=\linewidth, trim=0mm 2.5mm 0mm 2mm, clip]{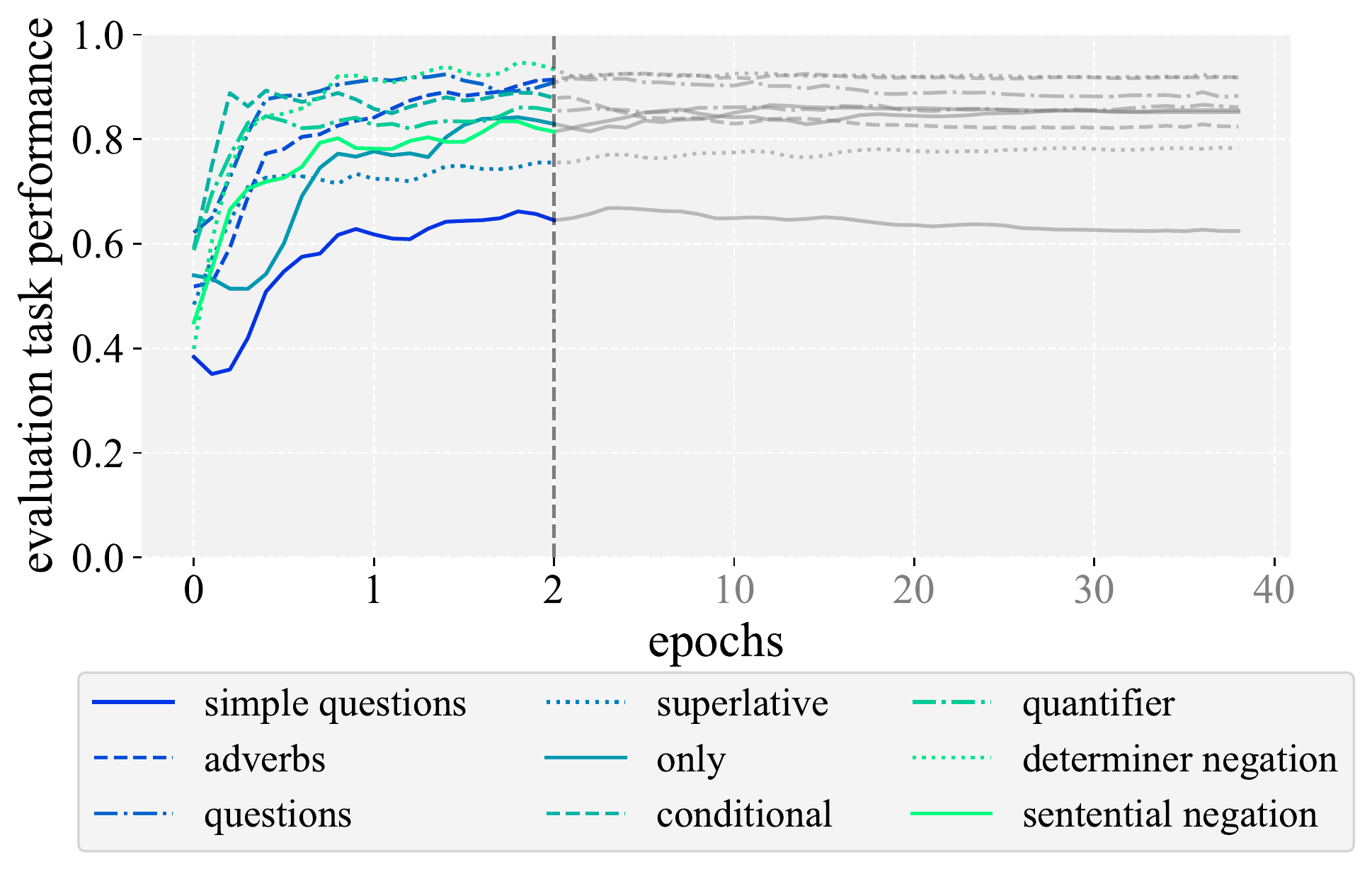}
     \vspace*{-2mm}
     \caption{Average evaluation task performance. The performance rises steeply during approximately the first 2 epochs of training and afterwards levels off.}
     \label{fig:trained_out}
\end{figure}

As a first step, we assess whether the LMs can adequately represent all nine categories of the evaluation task.
To do so, we train five models on the regular training corpus, and compute their final accuracy on our nine tasks.
All models show adequate performance on most contexts (see Table~\ref{tab:warstadt_accuracy}), with the exception of the simple question context.
Additionally, we observe that the models achieve their accuracy surprisingly fast: already after two epochs, there are no more substantial changes in empirical error (see Figure~\ref{fig:trained_out}).
In the rest of our experiments, we therefore focus only on these first two epochs.
\begin{table}[H]
    \centering
\begin{tabular}{lc} 
        \hline
        \textbf{Context} & \textbf{Accuracy} \scriptsize{$\pm$ std} \\
        \hline 
         Simple Questions & 0.62 \scriptsize{$\pm$ 0.05} \\
         Adverbs   & 0.92 \scriptsize{$\pm$ 0.01} \\ 
         Questions & 0.88 \scriptsize{$\pm$ 0.03} \\
         Superlative & 0.78 \scriptsize{$\pm$ 0.03} \\
         Only    & 0.86 \scriptsize{$\pm$ 0.04} \\
         Conditional & 0.82 \scriptsize{$\pm$ 0.06} \\
         Quantifier & 0.86 \scriptsize{$\pm$ 0.04} \\
         Determiner negation & 0.92 \scriptsize{$\pm$ 0.05} \\
         Sentential negation & 0.85 \scriptsize{$\pm$ 0.03} \\
        \hline      
\end{tabular}
    \caption{Performance of the LMs on the evaluation task after 40 epochs of training, averaged over 5 runs.}\label{tab:warstadt_accuracy}
\end{table}

\subsection{Frequency vs data efficiency}
\label{exp_1}
While some licensing contexts are rather common (e.g.\ negation), others appear scarcely as a licensor (e.g.\ adverbs). 
Therefore, throughout the learning process, the LMs encounter many instances of the more frequent contexts before they see an example of an infrequent context.
If LMs were able to leverage information across contexts, less frequent contexts should thus have more prior established NPI-understanding that they can bootstrap from. 
Consequently, the LMs should require fewer training examples to learn less frequent contexts than they need to learn more frequent contexts.
In other words, the LM should be more \emph{data efficient} for these infrequent contexts.

In our first experiment, we use this hypothesised relationship between frequency and data efficiency to assess whether LMs can exploit the similarities between different licensing contexts.
To be able to compare across different contexts, we quantify the data efficiency of an LM for a particular context as the number of examples the LM needs to observe until it reaches 95\% of its final accuracy for that context.\footnote{The \emph{more} data efficient, the \emph{lower} this number thus is.}
To make this measure more robust, we first apply a Savitzky–Golay noise-filter to the learning curve (degree of polynomial = 1, window size = 25; \citealt{Savitzky1964SmoothingProcedures}). \\[12pt]

\begin{figure}[H]
     \includegraphics[width=\linewidth, trim=0mm 3mm 0mm 4mm, clip]{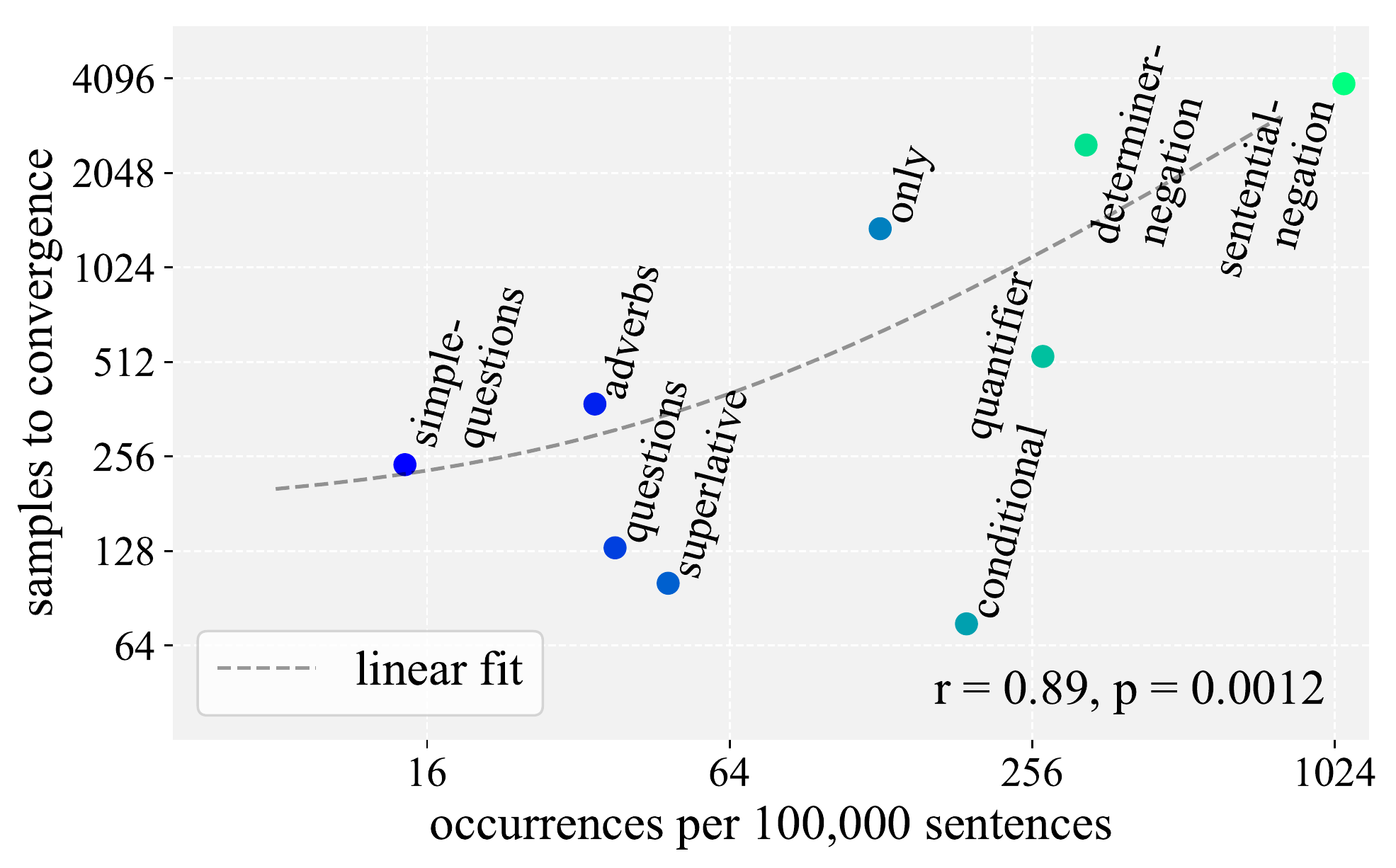}
     \caption{Data efficiency of nine different licensing contexts plotted against their frequency, averaged over five runs. The data efficiency is quantified as number of training examples the model needs to observe to achieve 95\% of the trained-out performance.}
     \label{fig:convergence_scatter}
\end{figure} 

We compute the data efficiency of the trained LMs for all nine contexts and compute the correlation between a context's frequency and the model's data efficiency with respect to that context.
In Figure~\ref{fig:convergence_scatter}, we plot the average data efficiency of each context against the frequency of that context, as well as the linear fit that relates these two variables.
The experiment demonstrates a strong relationship between the data efficiency and frequency of a respective context: \textit{r} = .89, p $<$ .05.
Hence, the less frequent a licensing context is, the fewer examples are needed for the model to learn it, from which we conclude that the model is indeed able to transfer knowledge from previously acquired knowledge.


\subsection{Transfer from general knowledge}
\label{exp_2}
While the presented relationship between frequency and data efficiency demonstrates that LMs can leverage previously learned information to learn less frequent licensing contexts, it does not unequivocally show that it leverages information from \emph{other NPI contexts}.
After all, when a less frequent context is encountered, the LM has not only had the opportunity to acquire prior knowledge about NPIs, it has also simply seen more language in general.
In other words, the LM may meanwhile also have acquired more \emph{general language knowledge}, which may help it to more quickly learn a less frequent licensing context.
In our second experiment, we isolate transfer from general language knowledge and transfer from previously observed NPIs by training LMs on \textit{single-context} corpora.

\begin{figure*}[!h]
     \centering
     \begin{subfigure}[b]{0.4\textwidth}
         \includegraphics[width=\textwidth, trim=0 5mm 0 2mm]{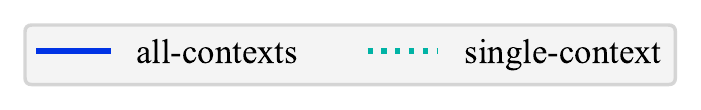}
         \label{fig:legend_fig_5}
     \end{subfigure}   
     
     \begin{subfigure}[b]{0.33\textwidth}
         \includegraphics[width=\textwidth, trim=0 1mm 0 2mm, clip]{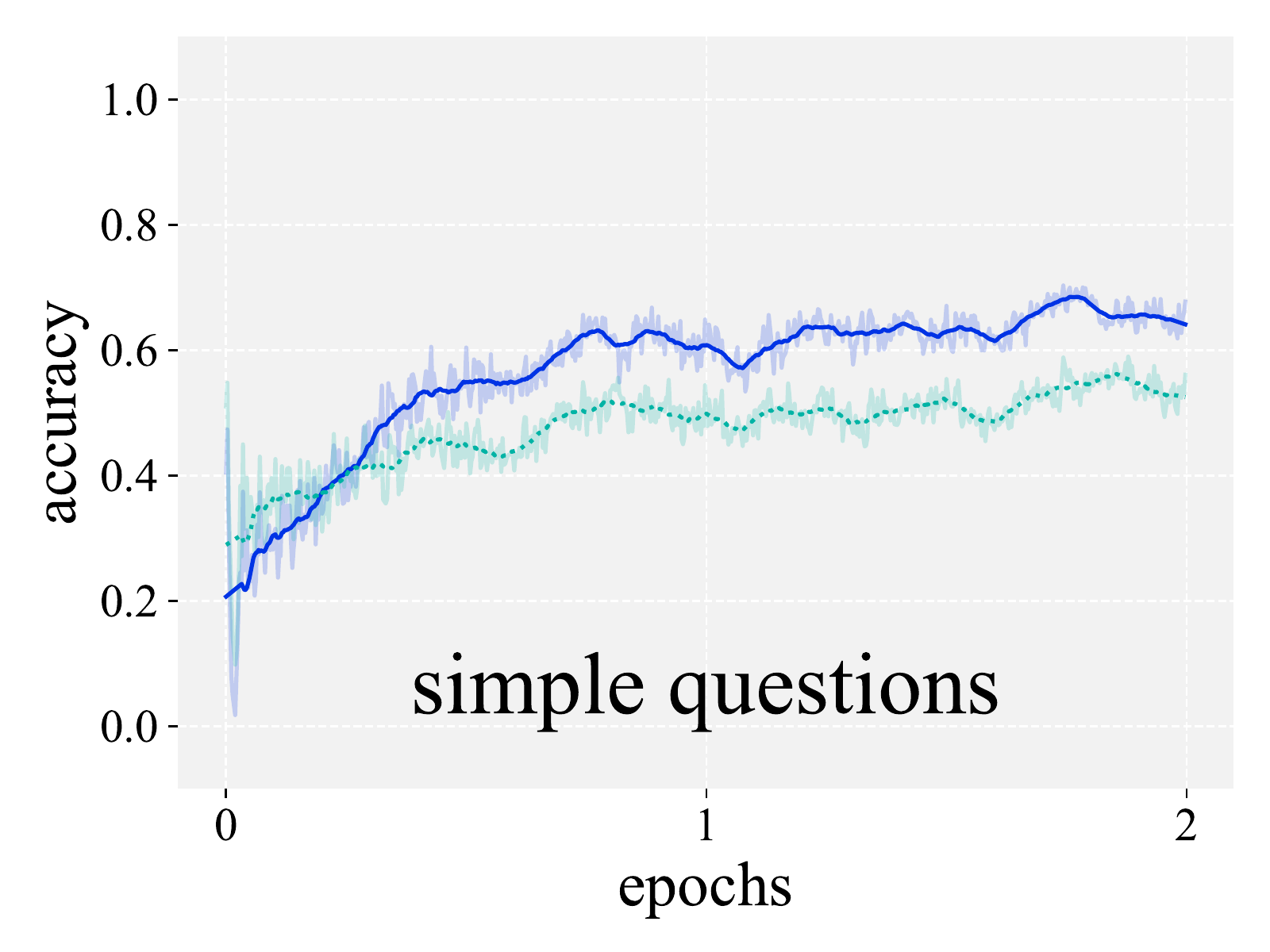}
         \label{fig:learning_curve_simplequestions}
     \end{subfigure}
     \hspace{-2mm}
     \begin{subfigure}[b]{0.33\textwidth}
         \includegraphics[width=\textwidth, trim=0mm 1mm 0 2mm, clip ]{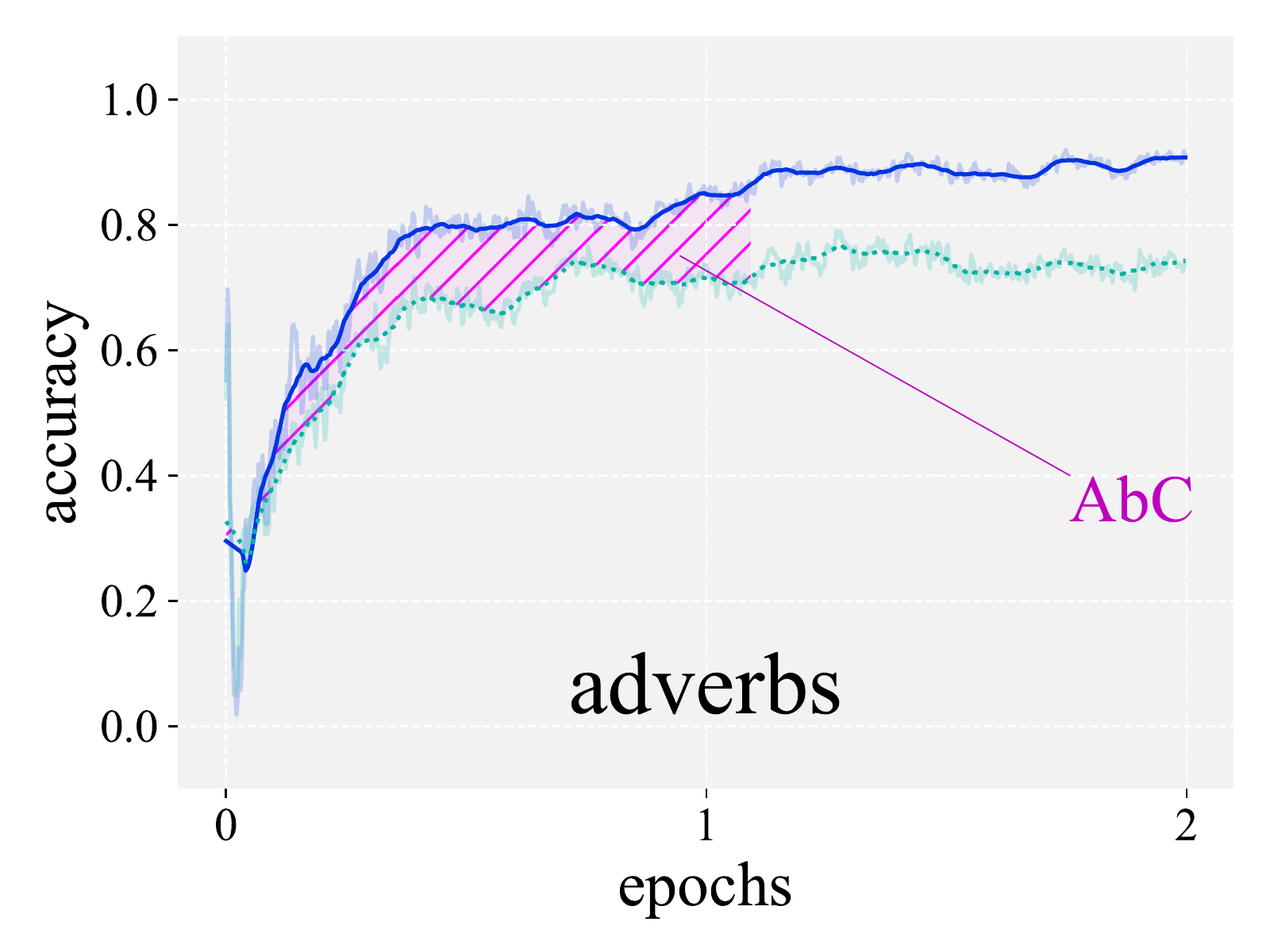}
         \label{fig:learning_curve_adverbs}
     \end{subfigure}
     \hspace{-2mm}
     \begin{subfigure}[b]{0.33\textwidth}
         \includegraphics[width=\textwidth, trim=0mm 1mm 0 2mm, clip]{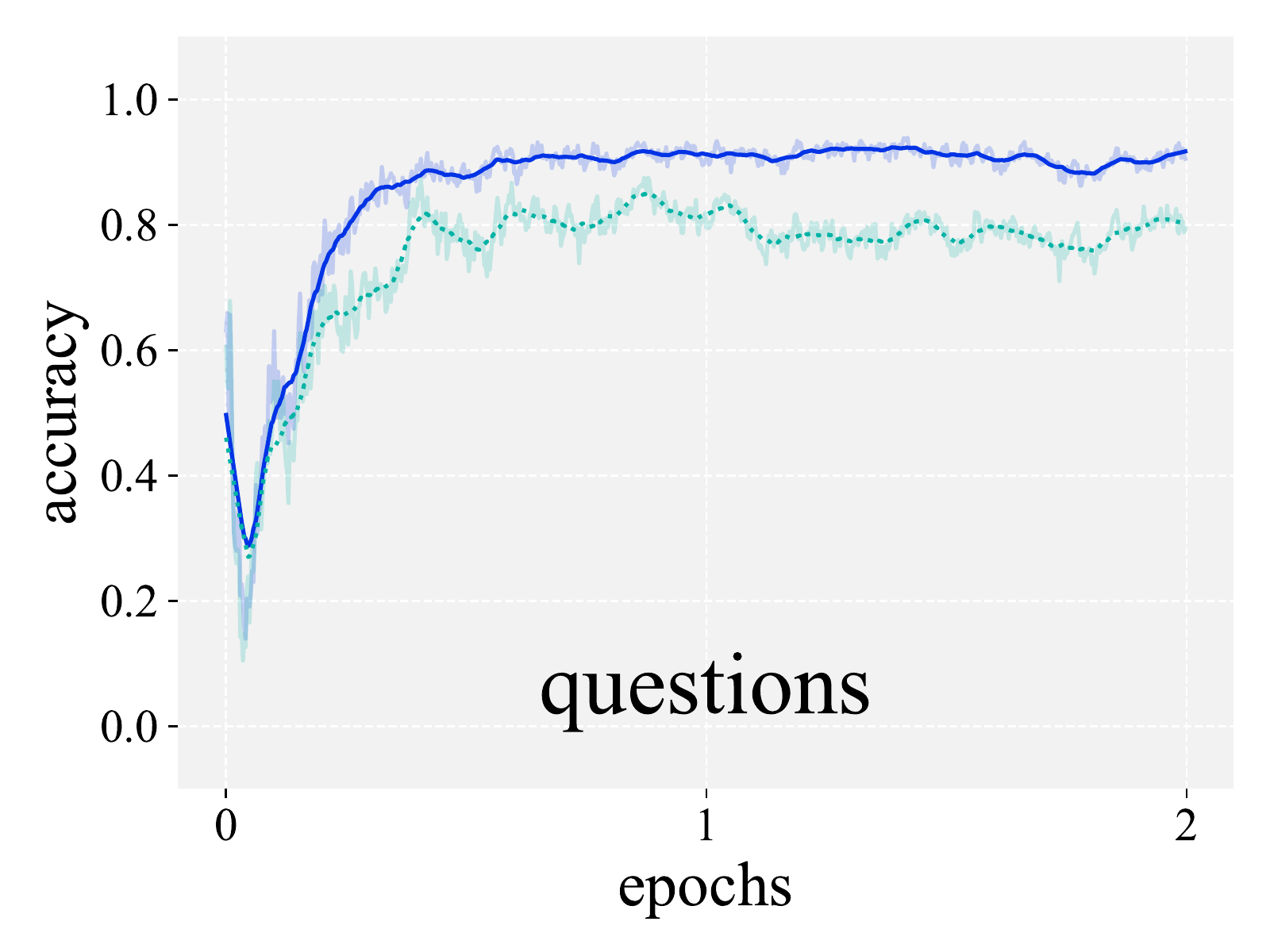}
         \label{fig:learning_curve_questions}
     \end{subfigure}\\
     \vspace{-5.7mm}
     \begin{subfigure}[b]{0.33\textwidth}
         \includegraphics[width=\textwidth, trim=0mm 1mm 0 2mm, clip]{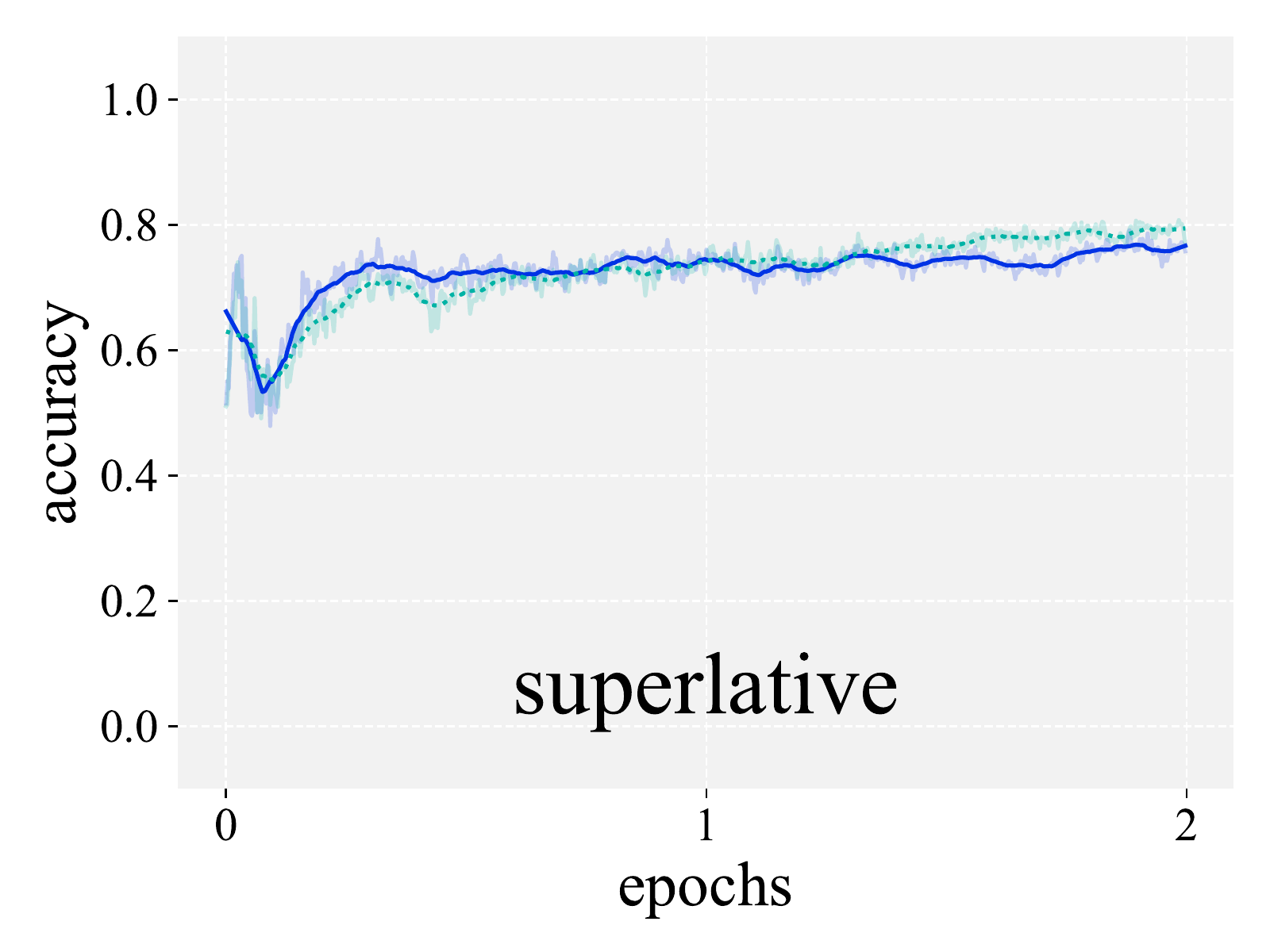}
         \label{fig:learning_curve_superlative}
     \end{subfigure} 
     \hspace{-2mm}
     \begin{subfigure}[b]{0.33\textwidth}
         \includegraphics[width=\textwidth, trim=0mm 1mm 0 2mm, clip]{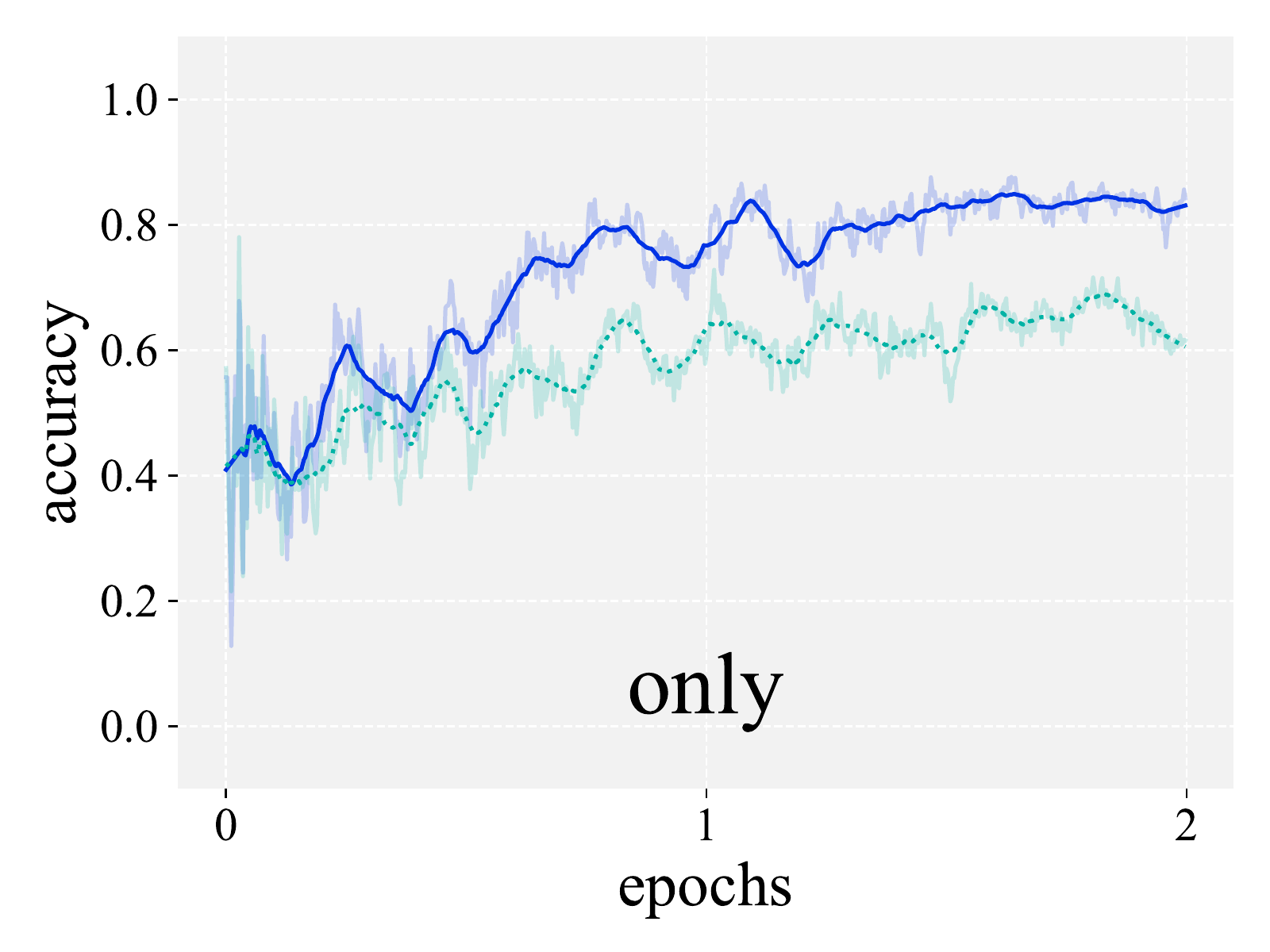}
         \label{fig:learning_curve_only}         
     \end{subfigure} 
     \hspace{-2mm}
     \begin{subfigure}[b]{0.33\textwidth}
         \includegraphics[width=\textwidth, trim=0mm 1mm 0 2mm, clip]{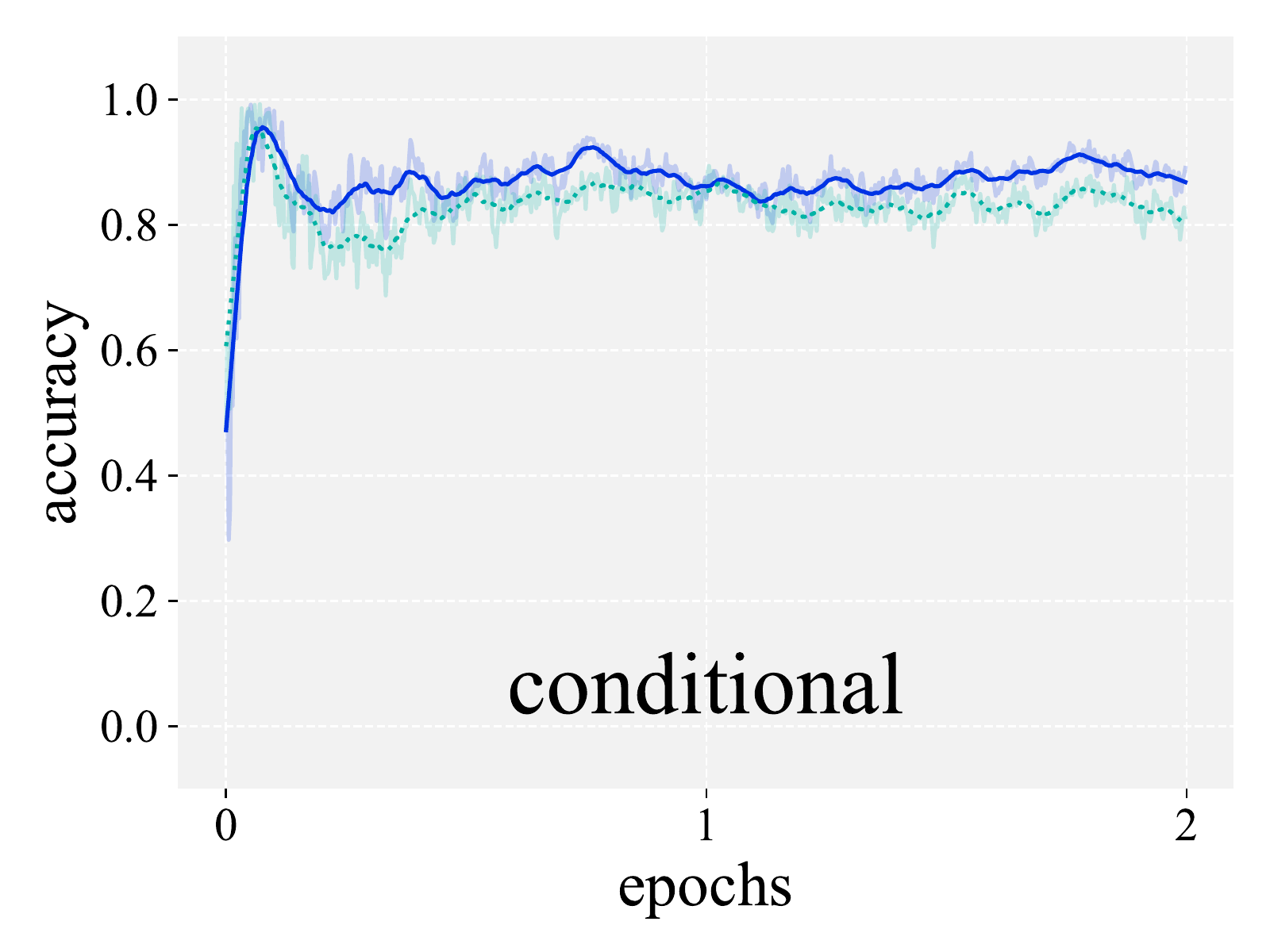}
         \label{fig:learning_curve_conditional}         
     \end{subfigure}\\
     \vspace{-5.7mm}
     
          \begin{subfigure}[b]{0.33\textwidth}
         \includegraphics[width=\textwidth, trim=0mm 1mm 0 2mm, clip]{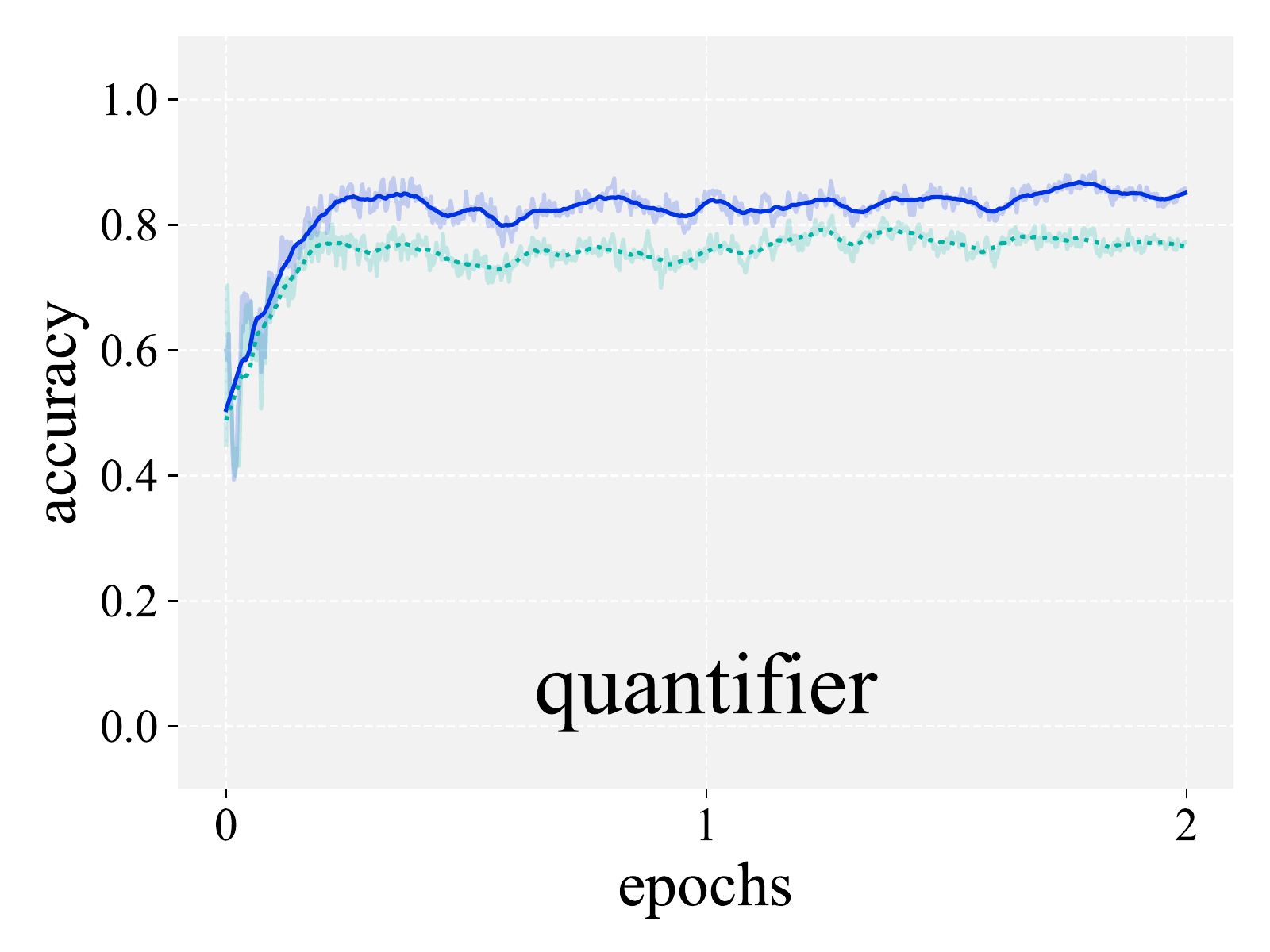}
         \label{fig:learning_curve_quantifier}         
     \end{subfigure} 
     \hspace{-2mm}
     \begin{subfigure}[b]{0.33\textwidth}
         \includegraphics[width=\textwidth, trim=0mm 1mm 0 2mm, clip]{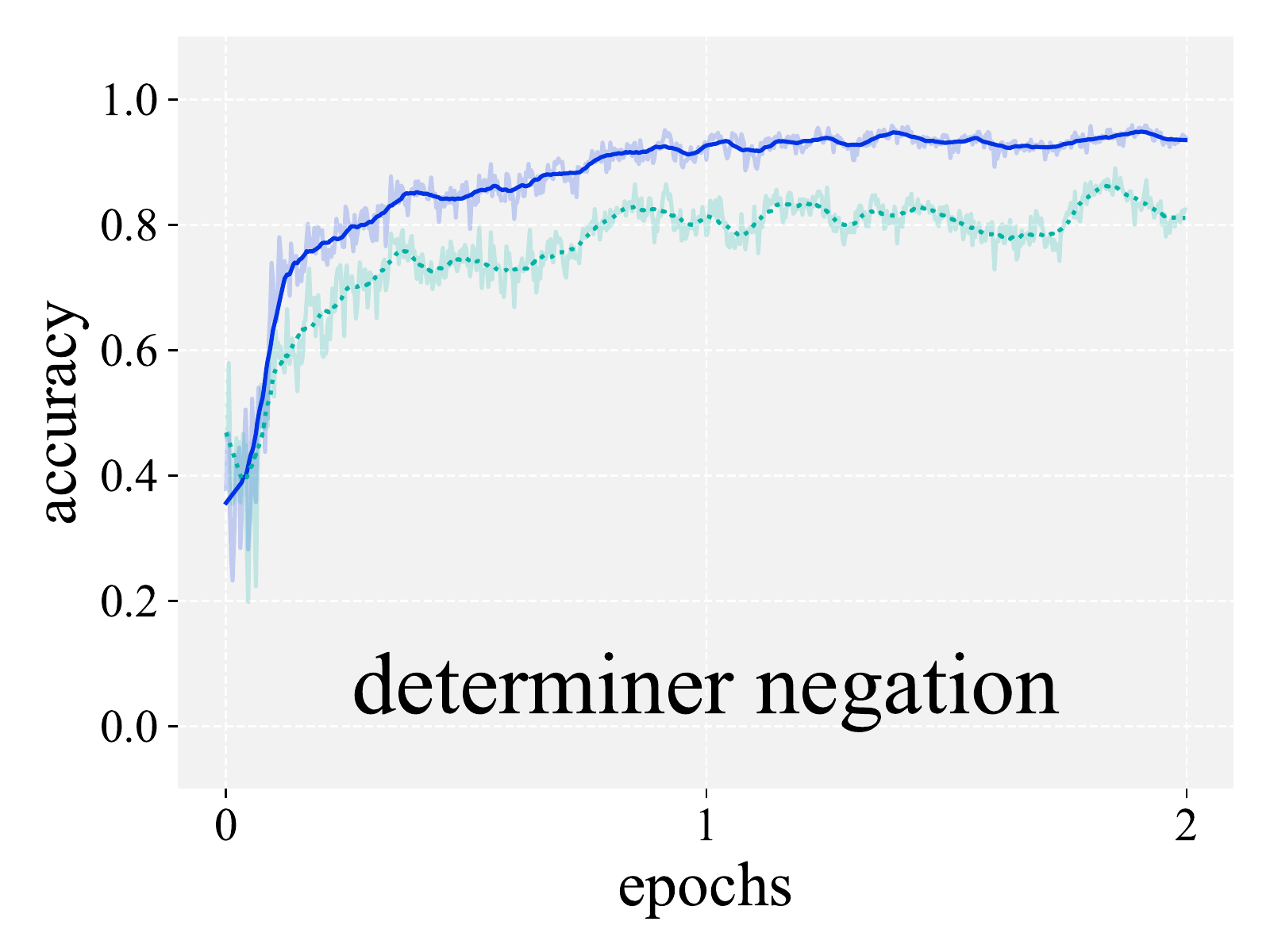}
         \label{fig:learning_curve_determiner_negation_biclausal}         
     \end{subfigure} 
     \hspace{-2mm}
     \begin{subfigure}[b]{0.33\textwidth}
         \includegraphics[width=\textwidth, trim=0mm 1mm 0 2mm, clip]{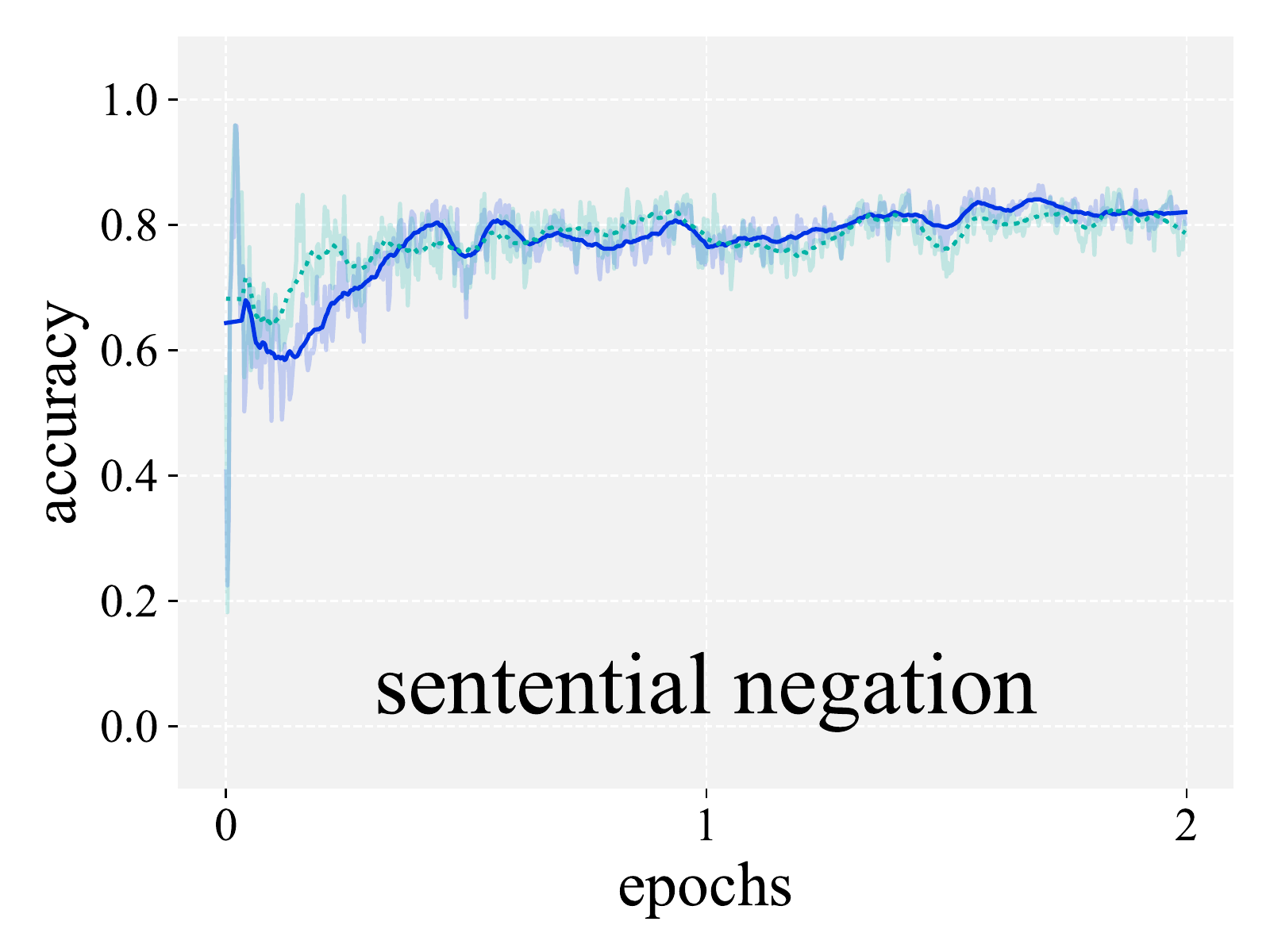}
         \label{fig:learning_curve_sentential_negation_biclausal}         
     \end{subfigure} 
     \vspace*{-5.7mm}
     \caption{The LMs performance on different licensing contexts for the first two epochs of training. We obtained these curves by evaluating all models at all 730 training-checkpoints on the evaluation task.}
     \label{fig:curves_all_contexts}
\end{figure*}

\paragraph{Single-context corpora}
\label{create_corpora}
\emph{Single-context corpora} contain NPIs licensed only by a single context.
LMs trained on these corpora can thus not transfer knowledge acquired from other licensing contexts, as these are not present in the training data.
By comparing the data efficiency of contexts between LMs trained on all-context and single-context corpora, we can thus infer how much of the increase of data efficiency for lower-frequent contexts is due to leveraging information from other contexts.

To create our nine single-context corpora, we use the procedure described in \S~\ref{filtering} to identify all sentences containing NPIs licensed by our nine contexts.
For every context, we then create a corpus in which all sentences containing other contexts licensing NPIs are replaced by a neutral sentence of the same length, sampled from the rest of the corpus.
During this replacement procedure, the ordering and composition of the corpus remained otherwise intact.

When we compare the learning of single-context with all-context models, we cannot rely on the previously used data-efficiency metric from Experiment~\ref{exp_1}. 
The data-efficiency measure is bound to how quickly the model reaches its final accuracy and accordingly benefits when its final accuracy decreases. 
As we expect the final accuracy to be lower in the single context models, comparing only data-efficiencies between models is likely to be uninformative.\footnote{Consider, for instance, the extreme case in which an LM does not learn a particular context at all anymore in the single-context condition, as indicated by a chance accuracy of 0.5. Because it is not learning anything, the model would arrive at its maximum accuracy before having seen any examples, resulting in a data efficiency of 0.}. 
In this experiment, as explained below, we instead consider the area between the curves (AbC).



\paragraph{Area between Curves (AbC)}
\label{AbC}
\emph{Area between Curves} (AbC) incorporates both data efficiency and accuracy: for every context, we calculate the area between the all-contexts and single-context learning curves until the point in time where they both have reached 95\% of their final accuracy.  
The larger this area is, the more impactful it is to remove all other NPI contexts, and the more the model leveraged from these contexts.
The learning curves of all contexts, along with an illustration of the AbC-measure, can be found in Figure~\ref{fig:curves_all_contexts}.

As a first interesting observation, we see that for seven of the nine contexts, the all-contexts model learns faster and achieves higher final performance.\footnote{A one-sided Welch's test confirms that the calculated AbCs are overall different from zero: \textit{t} = 2.61, \textit{p} $<$ .05.}
Both frequent and infrequent contexts thus benefit from information acquired by other licensing contexts, in terms of both data-efficiency and final accuracy.

\begin{figure}[H]
     \includegraphics[width=\columnwidth]{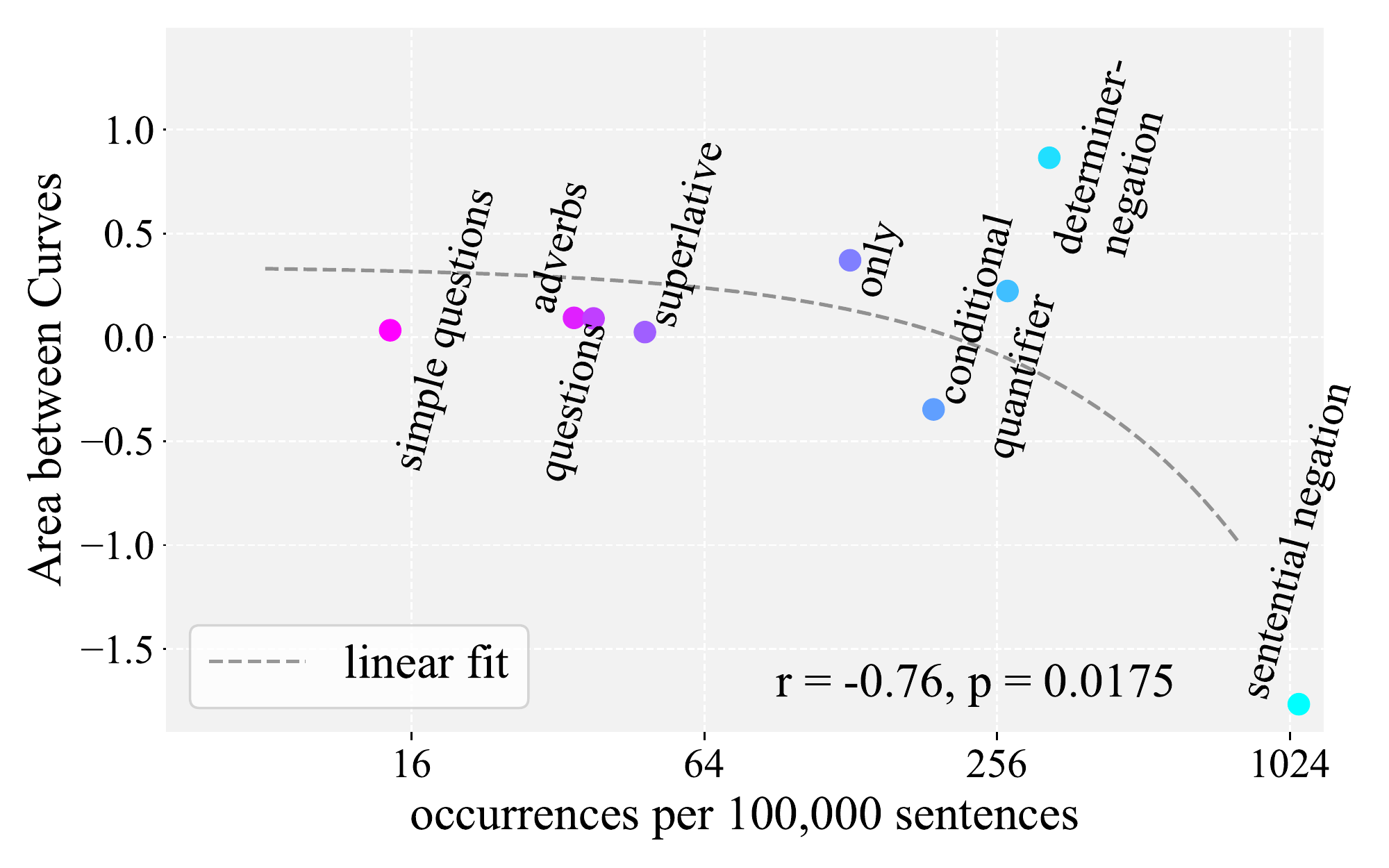}
     \caption{Normalised AbC for all licensing contexts until convergence of both contexts to 95\% accuracy. AbC $>$ 0 indicates a better performance of the all-context model and vice versa.}
     \label{fig:AbC_scatter}
\end{figure}

This positive transfer can also be seen in Figure~\ref{fig:AbC_scatter}, where we plot the AbC for all licensing contexts against their frequency.
This plot also confirms the relationship found in our previous experiments: the less frequent a context is, the more it benefits from other NPIs (\textit{r} = .76, \textit{p} $<$ .05).

\section{Discussion}
\label{sec:discussion}
In this paper, we studied language modelling as a multi-task problem.
We show that neural language models can find and exploit similarity between the different language construction rules that we deduced from linguistic theory and that their transfer behaviour mirrors the generalisation behaviour in traditionally constructed MTL settings.
In this section, we now reflect on how our setup and results contribute to the three different areas that we mentioned in the introduction: MTL, linguistics and interpretability research.

\subsection{Multi-task learning research}\label{discussion:MTL}
Studying LMs as multi-task learners, we observe several phenomena known from traditional MTL: when trained in parallel, similar (sub)tasks are learned more efficiently \citep[compare ][]{Collobert2011NaturalScratch, kaiser2017model}, and with higher accuracy \citep[][]{Collobert2008ALearning,kaiser2017model}, and this effect is stronger for less frequent tasks \citep[][]{benton-etal-2017-multitask, kaiser2017model}. 

Our study differs in one crucial aspect from previous research on MTL: it looks at learning dynamics \textit{within} one, larger, natural task instead of between tasks defined by the modeller.
As a consequence, the learning process itself is not constrained through a priori decisions concerning task selection, or how tasks should be optimised together.
In our scenario, contrary to traditional MTL, we use tasks and their hypothesised similarity only to \emph{analyse} the learning process of the language model, not to inform its training.
As such, our natural setting allows to study traditional MTL phenomena, such as data amplification, eavesdropping, and attention focusing, independent of arbitrary decisions regarding task selection and optimisation. 
This knowledge can then be transferred to scenarios in which more control over the selection of tasks may be required.

\subsection{Interpretability research}
\label{discussion:interpretability}
A second field where we believe studying language models as multi-task learners can contribute, is the field of interpretability.
On a more basic level, our paper confirms previous findings in interpretability that LMs are able to adequately model NPIs \citep{jumelet-hupkes-2018-language, Wilcox2019,Marvin2018TargetedModels}. 
We add to this literature by \textit{explicitly} showing that LMs are connecting different types of contexts together through their learning behaviour.  
Contrary to previous work, we are tapping the learning process itself as a source of information to better understand the inner workings of these models.

Traditional concepts from MTL, such as the earlier mentioned explanations of \citet{Caruana93multitasklearning:} and \citet{Ruder2017AnNetworks} (\S~\ref{subsec:multi_task_learning}) are valuable to better understanding what models are learning and how.
For instance, when we observe that the solution of models improves when more varied NPI material is presented (our single- versus all-context experiment), MTL can aid to formulate concrete hypotheses about \emph{why} this is the case.
This, in turn, can help us improve our understanding of the solutions that are learned by the model.
For instance, we find that the single-context models usually level-off on a lower accuracy-level than the all-context model (see Figure~\ref{fig:curves_all_contexts}). 
This is not merely explainable by the amount of data, as we continue to add training examples in either case. 
The difference between models instead appears to be due to the variety of the training data. 
The idea of \textit{attention focusing} \citep{Caruana93multitasklearning:,Caruana1997MultitaskLearning,Ruder2017AnNetworks} helps us to understand what is going on: by being trained on more varied NPI material, the model can better sort out which features are relevant and which ones are instead idiosyncrasies correlated with specific contexts.
Such hypotheses can then help inform further experiments, that investigate -- for example -- which features specifically are better learned through attention focusing.



\subsection{Linguistics research}
\label{discussion:linguistics}
Finally, 
we believe that studying language models as multi-task learners can also contribute to the field of linguistics.
In our study, we show that LMs can find and exploit similarities between linguistically defined concepts.
Turning things around, this generalisation behaviour of models can also be seen as a confirmation of the linguistic task hierarchy that we assumed from the start.
The language modelling objective is unconstrained by linguistic theory and therefore does not necessarily have to find the same solutions as linguistics. 
Similarity derived from the learning behaviour of language models might therefore be used as a tool to work on more disputed ideas in linguistics and to form new hypotheses in linguistic theory. 
While the linguistic insights that can be drawn from the current study are relatively limited, they do provide a proof of concept for future work: we show that domain knowledge and learning behaviour of neural models can be connected.



%

\section{Conclusion}
\label{sec:conclusion}
In the current study we explored the possibility to use multi-task learning as a framework to study learning behaviour \textit{within} a task.
To this end we considered LMs as multi-task learners and investigated how they learn the task-cluster of NPI-licensing. 
We find that LMs pick up on similarities that we assume from linguistic theory and exploit them to learn similar language constructions with less data and to a higher accuracy. 
Especially less frequent tasks benefit from this effect.

These results resemble positive transfer in `traditional' MTL. 
We lined out the possible benefits that our study may have for MTL research, interpretability and linguistics. 
From here there are many directions for future work: targeting less comprehensively researched areas in linguistics to add empirical data to otherwise usually theoretical linguistic discussions, investigating the change of internal representations in place of the behavioural measure used here to more precisely describe the learning process, or applying the approach to other high-level tasks in other modalities obeying other knowledge domains are just few of theses possibilities.

\section*{Acknowledgments}
We thank the anonymous reviewers; the COLT-group at UPF for the discussions and their useful feedback; the participants of the EviL-seminars, and especially Emmanuel Dupoux, for inspiration; and Laura Castro Moreno for her help with the graphic designs. Further, LW thanks the Department of Translation and Language Sciences at the University Pompeu Fabra for funding.

\typeout{} 

\newpage
\appendix




\onecolumn
\section{List of NPIs}
\label{sec:appendix_2}

We here present the full list of 160 NPIs that has been used for modifying the corpora:

\setlength{\columnsep}{-0.5mm}
\begin{multicols}{4}
\begin{itemize}\setlength{\itemsep}{-1mm}
    \item a bed of roses
    \item a care in the world
    \item a chance in hell
    \item a damn
    \item a damn thing
    \item a day goes by
    \item a day over
    \item a ghost of a
    \item a hair out of place
    \item a living soul
    \item a moment of your time
    \item a moment too soon
    \item a shadow of a doubt
    \item a single soul
    \item all that much
    \item all that many
    \item any
    \item any longer
    \item any old
    \item any time soon
    \item anybody
    \item anymore
    \item anyone
    \item anything
    \item anything like
    \item anytime soon
    \item anywhere
    \item anywhere close
    \item anywhere near
    \item as of yet
    \item as yet
    \item at all
    \item avail
    \item bat an eye
    \item be any time
    \item be anything like
    \item beat around the bush
    \item by a long sho
    \item by any chance
    \item by any means
    \item by any stretch
    \item by miles
    \item by much
    \item can be bothered
    \item can compare to
    \item can hold a candle to
    \item can make of
    \item can possibly
    \item chance in hell
    \item come at a worse time
    \item come cheap
    \item could care less
    \item could possibly
    \item cut the mustard
    \item even once
    \item ever
    \item far wrong
    \item for much longer
    \item for shit
    \item for the life of
    \item for the soul of
    \item give a crap
    \item give a damn
    \item give a fuck
    \item give a shit
    \item half a chance
    \item half bad
    \item have a clue
    \item have any of
    \item hold a candle to
    \item hold water
    \item in a blue moon
    \item in a hundred years
    \item in a long time
    \item in a million years
    \item in ages
    \item in all of history
    \item in any
    \item in any manner
    \item in any way
    \item in centuries
    \item in days
    \item in decades
    \item in his right mind
    \item in hours
    \item in living memory
    \item in minutes
    \item in months
    \item in recent memory
    \item in the least
    \item in the least bit
    \item in the slightest
    \item in weeks
    \item in years
    \item just any
    \item just yet
    \item know the first thing
    \item know the first thing about
    \item know the half of it
    \item least of all
    \item let alone
    \item lift a finger
    \item make a sound
    \item make head or tail of
    \item make much difference
    \item mean a thing
    \item mean feat
    \item miss a beat
    \item much care
    \item much help
    \item much of a
    \item much of anything
    \item much to look at
    \item much to lose
    \item nor
    \item on speaking terms
    \item on your life
    \item one single thing
    \item or anything
    \item rhyme or reason
    \item say much
    \item see eye to eye
    \item set foot
    \item set foot in
    \item set foot on
    \item sit right with
    \item sit well
    \item sit well with
    \item small feat
    \item so much as
    \item square with
    \item squat
    \item stand a chance
    \item strong suit
    \item such thing
    \item sweat it
    \item take his eyes off
    \item take kindly to
    \item take lightly
    \item take no for an answer
    \item that many
    \item that much
    \item that often
    \item the ghost of
    \item the half of
    \item the half of it
    \item the least bit
    \item the like of which
    \item the likes of which
    \item the slightest
    \item the slightest bit
    \item think much of
    \item to be taken lightly
    \item whatever
    \item whatsoever
    \item with a barge pole
    \item worth a damn
    \item worth his salt
    \item worth its salt
    \item yet
\end{itemize}
\end{multicols}


\end{document}